\documentclass[11pt]{article}
\usepackage[preprint]{acl}
\usepackage{times}
\usepackage{latexsym}
\usepackage{booktabs}
\usepackage{multirow}

\usepackage{fontawesome5}

\usepackage[T1]{fontenc}

\usepackage[utf8]{inputenc}

\usepackage[table]{xcolor}
\definecolor{heat}{HTML}{2E8B57}

\usepackage{array}

\usepackage{microtype}

\usepackage{inconsolata}

\usepackage{graphicx}

\title{Cross-Linguistic Effects in Bilingual Phoneme 
BabyLMs}

\author{
 \textbf{Nikitas Theodoropoulos\textsuperscript{1}},
 \textbf{Maria Lymperaiou\textsuperscript{2}},
 \textbf{Giorgos Filandrianos\textsuperscript{3}}
\\
\\
 \textsuperscript{1}Independent Researcher,\\
 \textsuperscript{2}National Technical University of Athens,\\
 \textsuperscript{3}Instituto de Telecomunicações, Portugal
\\
 \small{
   \textbf{Correspondence:} \href{mailto:email@domain}{nikitastheodorop@gmail.com}
 }
}

\begin{document}
\maketitle
\begin{abstract}
Cross-linguistic effects are a central topic in bilingual first-language acquisition. Artificial learners can help investigate L1--L2 interactions by enabling controlled comparisons across language combinations and learning conditions. Recent work explores this direction by training bilingual language models under developmentally plausible constraints. However, human and model learners still diverge in fundamental ways, with one major difference being input modality: children learn primarily from spoken input, whereas language models are typically trained on orthographic text. To reduce this gap, researchers have trained models on phonemic representations of speech. In this work, we combine these research directions to train bilingual BabyLMs with phonemic input.  We keep English fixed as the L2 and vary the L1 across German, Swedish, Persian, and Basque, selected to represent contrasting combinations of syntactic and phoneme-inventory distance from English. Our results show stronger L1-related variation in grammatical learning trajectories under phonemic than orthographic input, while early lexical differences align with phoneme-inventory similarity.
\\~\\
\setlength{\arrayrulewidth}{0.4pt}   
\setlength{\tabcolsep}{8pt}          
\renewcommand{\arraystretch}{}    
\begin{tabular}{ >{\centering\arraybackslash}m{3ex} | l }
  \includegraphics[height=2.5ex]{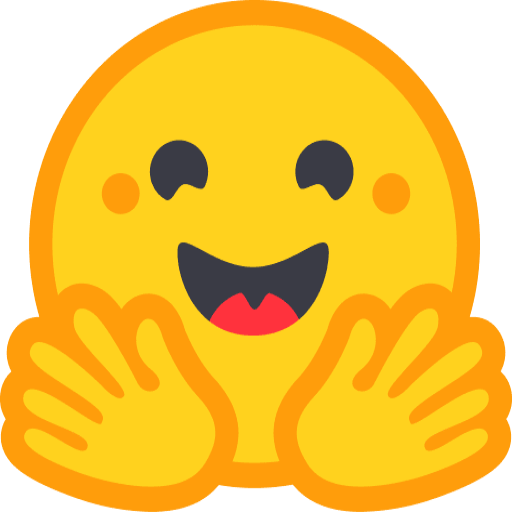}
    & \small \href{https://huggingface.co/collections/nikitastheo/bilingual-phone-lms}{nikitastheo/bilingual-phone-lms} \\
  \scalebox{1.2}{\faGithub}
    & \small \href{https://github.com/nikitas-theo/BilingualPhoneLMs}{nikitas-theo/BilingualPhoneLMs}\\
\end{tabular}
\end{abstract}

\section{Introduction}

Research in bilingual first-language acquisition has shown that children exposed to two languages early on develop two largely distinct linguistic systems. However, interactions between these systems can produce cross-linguistic effects \cite{Serratrice2013-tk}. Such effects are often asymmetric, with the first or more dominant language (L1) influencing the weaker or later-acquired language (L2), particularly at the level of morphosyntax. Despite extensive study of this topic, the exact conditions and strength of cross-linguistic influence remain unclear, with substantial variation across studies even for the same languages and phenomena \cite{VAN_Dijk2022-rt}. Addressing these questions requires controlled experiments across diverse language pairs and learning conditions.

\begin{figure}[t!]
    \centering
    \includegraphics[width=\linewidth]{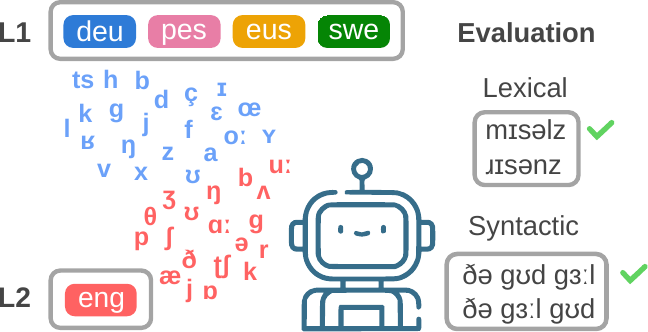}
    \caption{We train matched bilingual orthographic and phoneme LMs with four different L1s and English as the L2. We investigate L1-related differences in English grammatical performance with minimal-pair evaluations. Here, we illustrate the phoneme Syntactic and Lexical metrics from BabySLM \protect\cite{Lavechin_2023}.}
    \label{fig:intro}
\end{figure}

The BabyLM initiative \cite{conll-2023-babylm} supports the study of language acquisition through models trained under developmentally plausible settings that approximate the child's linguistic environment. Such artificial learners enable precise control over input and learning conditions, allowing the simulation of experiments that would otherwise be difficult or impossible with children \cite{warstadt_what_2022}. Recently this framework has been extended to bilingual acquisition to examine factors such as the effect of  bilingualism on learning \cite{zeng2026bringingbilingualbabylminvestigating}, critical period effects in models and humans \cite{constantinescu-etal-2025-investigating}, the role of child-directed speech \cite{binyamin-sulem-2026-learning}, and the importance of age of exposure \cite{issam2026languagemodelsartificiallearners}, among others.

Despite these advances, artificial learners remain limited by their reliance on orthographic input, whereas children acquire language primarily through speech. This mismatch matters because written and spoken representations encode different linguistic cues, and orthography may introduce patterns unavailable to children. Speech-based initiatives such as BabySLM \cite{Lavechin_2023} and BabyHuBERT \cite{charlot2026babyhubertmultilingualselfsupervisedlearning} offer more ecologically valid input, but suitable child-speech data are scarce and raise privacy and ethical challenges, especially in bilingual settings.  Phoneme-based language models provide a practical intermediate approach: they operate on phonemic representations that more closely approximate spoken input while retaining the efficiency, control, and data availability of text-based modeling.

In this work, we investigate whether cross-linguistic effects on L2 grammatical learning differ between bilingual LMs trained on orthographic and phonemic input. We use developmentally plausible multilingual corpora from BabyBabelLM \cite{jumelet-etal-2026-babybabellm}, to which we apply the G2P+ phonemization tool \cite{goriely-buttery-2025-ipa} to produce aligned phonemic corpora across languages. Specifically, we train matched orthographic and phoneme LMs, keeping English as the L2 and varying the L1 across German, Swedish, Persian, and Basque (Figure~\ref{fig:intro}). These languages represent different combinations of syntactic and phoneme-inventory distance from English, enabling us to examine the relationship between language transfer and specific typological distance measures. Concretely, we seek to answer the following questions: (a) Do cross-linguistic effects manifest differently according to modality (orthographic, phonemic)? (b) Does typological distance (syntactic, phoneme-inventory) correlate with modality-specific effects?

Following prior work on phoneme-based language models (e.g., \citet{bunzeck-etal-2025-small,goriely-etal-2024-babble}), we assess cross-linguistic effects through grammatical minimal-pair evaluations, extending our analysis of model behavior throughout  training. Our contributions are:

\begin{itemize}
\item We introduce a controlled framework for studying cross-linguistic influence in bilingual BabyLMs across orthographic and phonemic input modalities.
\item We train matched bilingual models with English as the L2 and four diverse L1s, covering distinct combinations of syntactic and phoneme-inventory distance.
\item We provide the first systematic analysis of whether L1-induced effects on L2 grammatical learning change when models are trained on phonemic rather than orthographic input.
\item We release the bilingual corpora in orthographic and phonemic form, the data-processing pipeline, trained models, and code to support further research on bilingual and phoneme-based artificial learners.
\end{itemize}

\section{Related work}

\paragraph{Cross-linguistic effects}
Research on bilingual development supports the emergence of independent linguistic systems, while showing that the two languages can influence one another selectively \cite{ParadisGenesee1996,Serratrice2013-tk}. These effects may be quantitative, facilitating or delaying the acquisition of a linguistic structure, or qualitative, producing patterns not observed in monolingual peers. Their occurrence has been linked to structural overlap and the syntax--pragmatics interface \cite{HulkMuller2000}. A large meta-analysis reports reliable small-to-moderate effects overall, while highlighting substantial variation across studies and identifying significant influence from the societal language to the non-societal language  \cite{VAN_Dijk2022-rt}. Cross-linguistic influence also extends to phonology: \citet{Paradis2001} finds that bilingual children develop separate but non-autonomous phonological systems. Related effects also arise in multilingual language models, where shared training can result in positive or negative interference depending on factors such as data balance, model capacity, and linguistic similarity \cite{wang-etal-2020-negative,chang-etal-2024-multilinguality} with syntactic similarity predicting greater benefits. We extend these research directions by investigating cross-linguistic effects in bilingual phoneme LMs and the role of phonemic input in shaping L1--L2 transfer.

\paragraph{Bilingual BabyLMs} There has been considerable research effort in modeling bilingual language acquisition with BabyLMs. \citet{zeng2026bringingbilingualbabylminvestigating} investigate whether bilingualism can negatively or positively influence language learning across various input conditions and scales, concluding no strong effect exists. \citet{issam2026languagemodelsartificiallearners} vary the onset of exposure to the second language, finding that both language dominance and similarity contribute to cross linguistic effects.  \citet{binyamin-sulem-2026-learning} analyze the effect of child-directed speech on semantic and syntactic knowledge. In second language acquisition, studies have focused on the effect of typological similarity on morphosyntactic performance \cite{shen_bambino-lm_2024, oba_second_2023, yadavalli-etal-2023-slabert}, predictive models of non-native human sentence processing \cite{aoyama_modeling_2024}, alignment with learner error patterns for different L1 \cite{gao-etal-2025-bliss}, interaction-driven continual pre-training \citet{shen_bambino-lm_2024}, and traces of L1 influence in L2 production \cite{barbenel-etal-2026-l1}. Our work extends these approaches by analyzing L1-induced effects across two input modalities: orthographic and phonemic input.

\paragraph{Phoneme BabyLMs}
Phonemes have been recently explored as a linguistically grounded input representation for developmentally plausible language models. Grapheme- and phoneme-based BabyLMs can acquire substantial linguistic abilities, with phoneme models approaching grapheme-based performance on various evaluations \cite{bunzeck-etal-2024-graphemes,bunzeck-etal-2025-small}. \citet{goriely-etal-2024-babble} compare orthographic and character-level input with continuous streams of phonemes, finding a modest performance reduction on conventional tasks but advantages for phonological analysis. \citet{goriely-buttery-2025-ipa} introduce G2P+ and IPA CHILDES, a phonemic child-centered speech resource covering 31 languages and show that phoneme LMs learn phonological features cross-lingually. It is further demonstrated that such models implicitly track word boundaries across languages \cite{goriely-buttery-2025-babylms}. We extend this research to bilingual acquisition by comparing L1-induced effects in matched bilingual orthographic and phoneme LMs.

\section{Methods}
To examine whether the patterns of cross-linguistic influence identified in prior work persist under a more speech-like representation, we compare bilingual orthographic and phoneme LMs.
Here, we define a cross-linguistic effect as the difference in model linguistic ability in the L2 that results from the influence of the L1. Given that monolingual children (and models) receive substantially more input in a language compared with bilingual peers, some performance differences are generally expected. Our main goal is thus to compare bilingual learners across different L1s, measuring the influence of input modality and typological distance on the acquisition of L2, while keeping monolingual models as an informative baseline. 

Orthographic LMs receive BPE-tokenized written input, whereas the phoneme LMs receive sequences of individual phoneme tokens. We adopt a controlled experimental design in which the same L1--L2 language pairs are used across both conditions. The transformer architecture, source data, training schedule, and optimization settings are held constant where possible, while the modality-specific tokenizers result in different vocabulary and embedding sizes. This setup allows us to isolate how input modality affects the influence of the L1 on L2 learning. We operationalize the L1--L2 relationship using two metrics: syntactic and phoneme-inventory distance. Concretely, we seek to uncover interactions between L1--L2 typological distance and L2 grammatical performance.

\subsection{Simulating bilingual acquisition}

To model  bilingual acquisition, we train GPT-2 models under two matched conditions: orthographic input and phonemic input. We refer to the resulting systems as \textbf{orthographic LMs} and \textbf{phoneme LMs}, respectively. The models share the same transformer architecture, with 85M non-embedding parameters, while the embedding layers differ in size because of the substantially different vocabularies: approximately 500k parameters for the phoneme and 23M for the orthographic LMs. As a first study of bilingual phoneme models, we opted to compare with BPE-trained orthographic models that constitute the prevailing bilingual learners used in computational works. Grapheme LMs (e.g., \citet{bunzeck-etal-2024-graphemes, goriely-etal-2024-babble}) trained on individual characters provide a better matched condition with respect to model capacity but are more rare in the literature, and no grapheme-based bilingual BabyLMs exist to date. This would make it hard to compare our models with prior work. Our framework can readily accommodate such learners in future research experiments.

In the phonemic condition, each phoneme is represented by an individual phoneme token, with dedicated tokens marking word and utterance boundaries. The phoneme-token vocabulary is constructed from the phonemic symbols observed in the corresponding bilingual training corpus. In the orthographic condition, we train a separate BPE tokenizer for each language pair, with a vocabulary of 30,000 tokens. For the phoneme LMs, the vocabulary consists of phonemes occurring more than 10 times in the corresponding bilingual corpus.

We draw the training data from BabyBabelLM \cite{jumelet-etal-2026-babybabellm}, a multilingual corpus designed for developmentally plausible language modeling. For each bilingual setting, we sample 10M words from the L1 and 10M words from English, for a total of 20M words. To ensure comparability across languages, sampling is calibrated using byte premiums, which account for language-specific differences in the relationship between word counts and encoded data size \cite{arnett-etal-2024-bit}, resulting in content-matched training data. Details about data distributions across languages are given in Appendix~\ref{app:data_distribution}. We then phonemize the same corpora with G2P+ \cite{goriely-buttery-2025-ipa}, a tool developed as a unified phonemization backend for languages in CHILDES, with aligned phonemic inventories across languages. This way, we produce paired orthographic and phonemic datasets from identical source utterances.

Following prior work on bilingual LMs \cite{arnett-etal-2025-acquisition,constantinescu-etal-2025-investigating}, we adopt a sequential-interleaved training regime. Models are first exposed exclusively to the L1 for 10k steps, after which they receive interleaved L1 and English input for a further 20k steps. Thus, our models correspond to early sequential bilinguals. We adopt the training configuration of the Goldfish model suite \cite{chang-etal-2026-goldfish}; full hyperparameters are reported in App.~\ref{app:training}. For each language combination, we train two bilingual LMs: one orthographic and one phoneme. As monolingual references, we additionally train one English orthographic and one English phoneme LM. For training, we use the 10M English portion in the bilingual corpora, which we supplement with 10M additional words from BabyBabelLM. These models are trained for 30k steps on the combined 20M corpus, matching the total L1--L2 exposure and data of the bilingual learners.

To increase the robustness of our results, for each model we report performance across 3 different random seeds. For each seed, we re-sample and phonemize all language data from BabyBabelLM to 10M words--leading to distinct but comparable corpora--and train orthographic and phoneme LMs. In the following results, when possible along with the mean over seeds we report standard deviation.

\subsection{Language selection}

\begin{figure}[t!]
    \centering
    \includegraphics[width=0.9\linewidth]{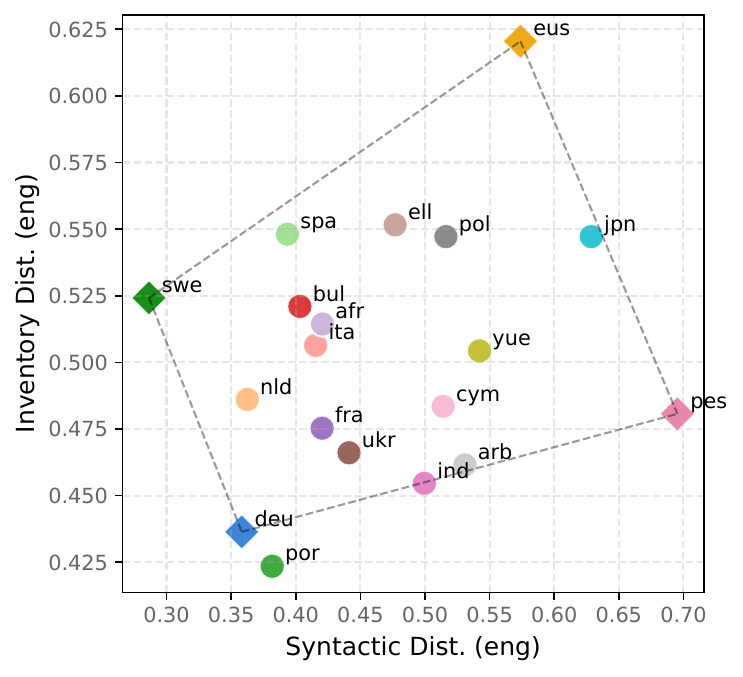}
    \caption{Syntactic and phoneme-inventory distances from English for the eligible language set, computed with URIEL+ \protect\cite{khan-etal-2025-uriel}. Languages selected for training the bilingual models are marked with $\Diamond$. We opt for languages that occupy contrasting regions of the syntactic and phoneme-inventory distance space.}
    \label{fig:uriel_distances}
\end{figure}

Language selection is central to our design because the effect of an L1 on English may depend on typological similarity. Syntactic measures of similarity have been shown to be predictive of transfer effects \cite{chang-etal-2024-multilinguality}. Given that we evaluate on grammatical benchmarks, we also expect an influence  in our models. We additionally consider phoneme-inventory distance to capture differences in the phonemic inventories of the two languages. Therefore, we select L1s that vary along these two dimensions, enabling us to examine whether the patterns observed under orthographic and phonemic input are associated with typological measures.

We keep English fixed as the L2 to facilitate comparison with prior work and because suitable evaluation resources are more widely available for English in both orthographic and phonemic form. Candidate L1s are restricted to languages in BabyBabelLM with at least 10M words of training data, ensuring that all models can be trained under the same developmentally plausible data constraints.

For each candidate language, we compute its syntactic and phoneme-inventory distance from English using URIEL+ \cite{khan-etal-2025-uriel}. Syntactic distance captures differences in typological syntactic grammatical features, whereas phoneme-inventory distance reflects differences between the sets of phonemes used by the two languages\footnote{URIEL+ also provides morphological and phonological distance measures, but their coverage was insufficient for the full candidate set.}. These complementary measures are used to characterize syntactic and phoneme-inventory similarity between each L1 and English respectively. Distances are illustrated in Figure~\ref{fig:uriel_distances}.

We select German (\texttt{deu}), Swedish (\texttt{swe}), Persian (\texttt{pes}), and Basque (\texttt{eus}) to represent four contrasting regions of the resulting two-dimensional space: low syntactic and low phoneme-inventory distance for German, low syntactic and high phoneme-inventory distance for Swedish, high syntactic and low phoneme-inventory distance for Persian, and high syntactic and high phoneme-inventory distance for Basque. This contrastive selection samples four regions of the distance space, allowing us to explore whether performance patterns align with syntactic or phoneme-inventory similarity. We hypothesize that syntactic distance will be more closely associated with grammatical evaluation performance, whereas phoneme-inventory distance will be more informative for phonologically oriented evaluations and may exert a stronger influence under phonemic input.

\begin{figure*}[!t]
    \centering
    \includegraphics[width=\linewidth]{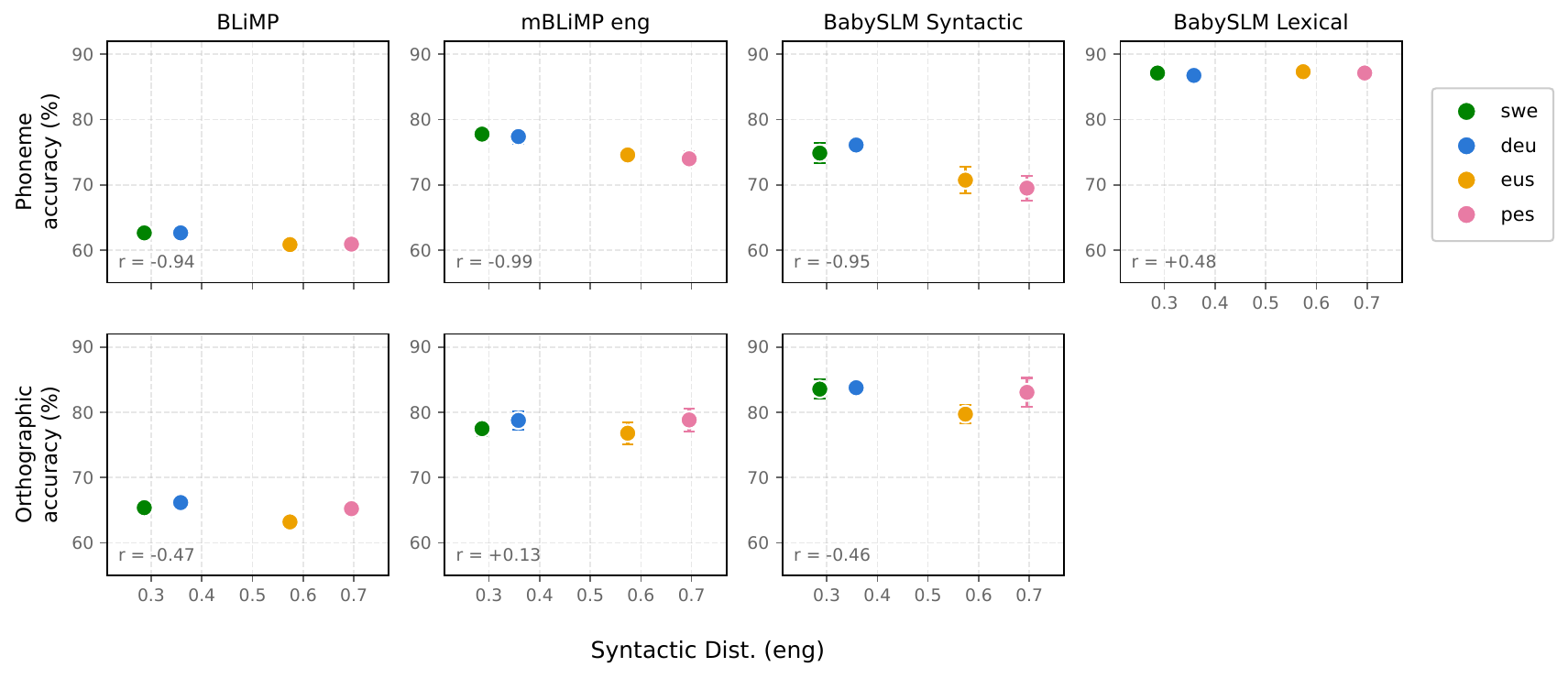}
    \caption{English evaluation accuracy as a function of L1--English syntactic distance for phoneme (top) and orthographic (bottom) LMs. For each model we report an average across 3 seeds with SD error bars. Pearson correlation ($r$) over averages shown for each benchmark. BabySLM Lexical is available only in phonemic form.}
    \label{fig:acc_dots}
\end{figure*}

\subsection{Evaluation}

We use minimal-pair tasks to assess English grammatical learning, targeted grammatical knowledge in both languages, and the effect of input modality. A model receives credit when it assigns higher score to the grammatical sentence, where sentence scores are computed as the sum of token log probabilities. We use the latest BabyLM evaluation pipeline \footnote{\url{https://github.com/babylm-org/babylm-eval}} to evaluate on three benchmarks.

BLiMP introduced by \citet{warstadt-etal-2020-blimp-benchmark}, comprises 67 minimal-pair datasets across syntactic, morphological, and semantic phenomena. Since English is fixed as the L2, variation across L1 conditions provides our main measure of L1-related effects on L2 grammatical learning. We additionally use MultiBLiMP \cite{jumelet-etal-2026-multiblimp}, which covers two subject--verb agreement phenomena in 101 languages, as a targeted check of grammatical learning in both English and each L1.

For the phoneme LMs, we convert both benchmarks into phonemic representations using G2P+ \cite{goriely-buttery-2025-ipa}, following the same preprocessing used for the training data. We also report the BabySLM syntactic and lexical evaluations \cite{Lavechin_2023}: the former tests six simple grammatical phenomena, while the latter contrasts real words with phonotactically plausible pseudowords and provides a targeted subword diagnostic for phoneme LMs.

\section{Results}

\subsection{Minimal-Pair Evaluation}

Table~\ref{tab:results} reports minimal-pair accuracy for the bilingual orthographic and phoneme LMs across all L1 conditions, together with the corresponding English monolingual references. We first examine each input condition separately and then compare the extent and direction of L1-related variation across modalities. Overall, performance across seeds stays consistent, with occasional variations.

\paragraph{Orthographic LMs}
The English monolingual reference achieves the highest score on all three applicable evaluations: BLiMP, English MultiBLiMP, and BabySLM Syntactic. The bilingual models nevertheless perform above chance on BLiMP and on the English and L1 portions of MultiBLiMP, indicating that they acquire at least some of the evaluated grammatical contrasts in both languages.

\begin{table}[!t]
\centering
\resizebox{\linewidth}{!}{
\begin{tabular}{ll|ccccc}
\toprule
 &  &  & \multicolumn{2}{c}{mBLiMP} & \multicolumn{2}{c}{BabySLM} \\
\cmidrule(lr){4-5} \cmidrule(lr){6-7}
 & \textbf{L1} & BLiMP & L1 & eng & Lexical & Syntactic \\
\midrule\addlinespace[2.5pt]
\multirow{5}{*}{\rotatebox{90}{\textbf{Orthographic}}} & eng & \cellcolor{heat!55} 69.1 & -- & \cellcolor{heat!42} 83.4 & -- & \cellcolor{heat!53} 88.2 \\
 & deu & \cellcolor{heat!36} 66.2 & \cellcolor{heat!2} 84.7 & \cellcolor{heat!21} 78.8 & -- & \cellcolor{heat!40} 83.8 \\
 & eus & \cellcolor{heat!16} 63.2 & \cellcolor{heat!18} 87.2 & \cellcolor{heat!12} 76.8 & -- & \cellcolor{heat!29} 79.7 \\
 & pes & \cellcolor{heat!29} 65.2 & \cellcolor{heat!1} 84.6 & \cellcolor{heat!21} 78.8 & -- & \cellcolor{heat!38} 83.1 \\
 & swe & \cellcolor{heat!30} 65.4 & -- & \cellcolor{heat!16} 77.5 & -- & \cellcolor{heat!39} 83.5 \\
\midrule
\multirow{5}{*}{\rotatebox{90}{\textbf{Phoneme}}} & eng & \cellcolor{heat!37} 66.4 & -- & \cellcolor{heat!55} 86.4 & \cellcolor{heat!55} 88.6 & \cellcolor{heat!55} 89.0 \\
 & deu & \cellcolor{heat!12} 62.6 & \cellcolor{heat!0} 84.4 & \cellcolor{heat!15} 77.4 & \cellcolor{heat!0} 86.8 & \cellcolor{heat!19} 76.1 \\
 & eus & \cellcolor{heat!0} 60.8 & \cellcolor{heat!55} 93.0 & \cellcolor{heat!3} 74.6 & \cellcolor{heat!15} 87.3 & \cellcolor{heat!3} 70.7 \\
 & pes & \cellcolor{heat!1} 60.9 & \cellcolor{heat!17} 87.0 & \cellcolor{heat!0} 74.0 & \cellcolor{heat!9} 87.1 & \cellcolor{heat!0} 69.5 \\
 & swe & \cellcolor{heat!12} 62.6 & -- & \cellcolor{heat!17} 77.8 & \cellcolor{heat!9} 87.1 & \cellcolor{heat!15} 74.9 \\
\bottomrule
\end{tabular}

}
\caption{Minimal-pair accuracy for the English monolingual references and bilingual LMs trained on orthographic or phonemic input. Results are averaged across 3 different seeds. BLiMP and MultiBLiMP (mBLiMP) are evaluated in their original orthographic form for the orthographic LMs and in phonemic form for the phoneme LMs. BabySLM Lexical is available only for the phoneme condition.}
\label{tab:results}
\end{table}

Based on prior findings that syntactic similarity can facilitate multilingual transfer \cite{chang-etal-2024-multilinguality}, we expected the German--English and Swedish--English models, whose L1s are syntactically closer to English, to achieve higher English grammatical performance. We observe some evidence for this pattern in the orthographic condition. German and Swedish obtain relatively high scores on BLiMP and BabySLM Syntactic, while the more syntactically distant Basque model performs lowest on both benchmarks. However, the pattern is not fully consistent: Persian performs similarly to German and Swedish on BabySLM Syntactic and BLiMP, and German and Persian obtain the same rounded score on English MultiBLiMP. Accordingly, syntactic distance shows only a weak and benchmark-dependent association with grammatical performance for orthographic LMs. This is broadly consistent with \citet{zeng2026bringingbilingualbabylminvestigating}, who find no large or uniform effects of bilingual training under orthographic input.

\paragraph{Phoneme LMs}

The English reference achieves the highest absolute performance across all applicable evaluations. Importantly, we notice it also surpasses the matched orthographic English LM in two out of the three common evaluations (BabySLM Syntactic, mBLiMP eng), consistent with previous findings of increased performance for phoneme LMs in some cases \cite{bunzeck-etal-2025-small, goriely-etal-2024-babble}. For bilingual LMs we notice overall matched or reduced performance when compared with orthographic counterparts.

However, a clearer L1-related ordering emerges across the grammatical benchmarks for bilingual phoneme LMs. German and Swedish, the two L1s with lower syntactic distance from English, consistently outperform Persian and Basque on BLiMP, English MultiBLiMP, and BabySLM Syntactic.

The ordering among bilingual models is more consistent across the grammatical evaluations. German achieves the highest BabySLM Syntactic scores, while German and Swedish obtain similar scores for English BLiMP and MultiBLiMP. Persian and Basque generally obtain lower grammatical scores. This pattern aligns with syntactic distance: German and Swedish are syntactically closer to English than Persian and Basque. Because German, Swedish, and English are also genealogically related, the observed grouping may reflect a combination of syntactic similarity, lexical overlap, and other shared properties rather than the URIEL+ syntactic measure alone. Persian is particularly informative because it has relatively low phoneme-inventory distance but high syntactic distance from English. Its lower grammatical scores indicate that phoneme-inventory similarity alone is not sufficient to predict performance on these grammatical evaluations. However, because each combination of syntactic and phoneme-inventory distance is represented by only one language, the observed ordering cannot be attributed independently to syntactic distance rather than to other language-specific properties.

The BabySLM lexical task exhibits little variation at the final checkpoint: all bilingual phoneme LMs score between 86.8 and 87.3, close to the monolingual score of 88.6. Final accuracy therefore provides no evidence of a persistent L1-related difference on this task. As shown by the checkpoint analysis, however, this convergence conceals differences that arise earlier in training.

\begin{figure*}[h!]
    \centering
    \includegraphics[width=\linewidth]{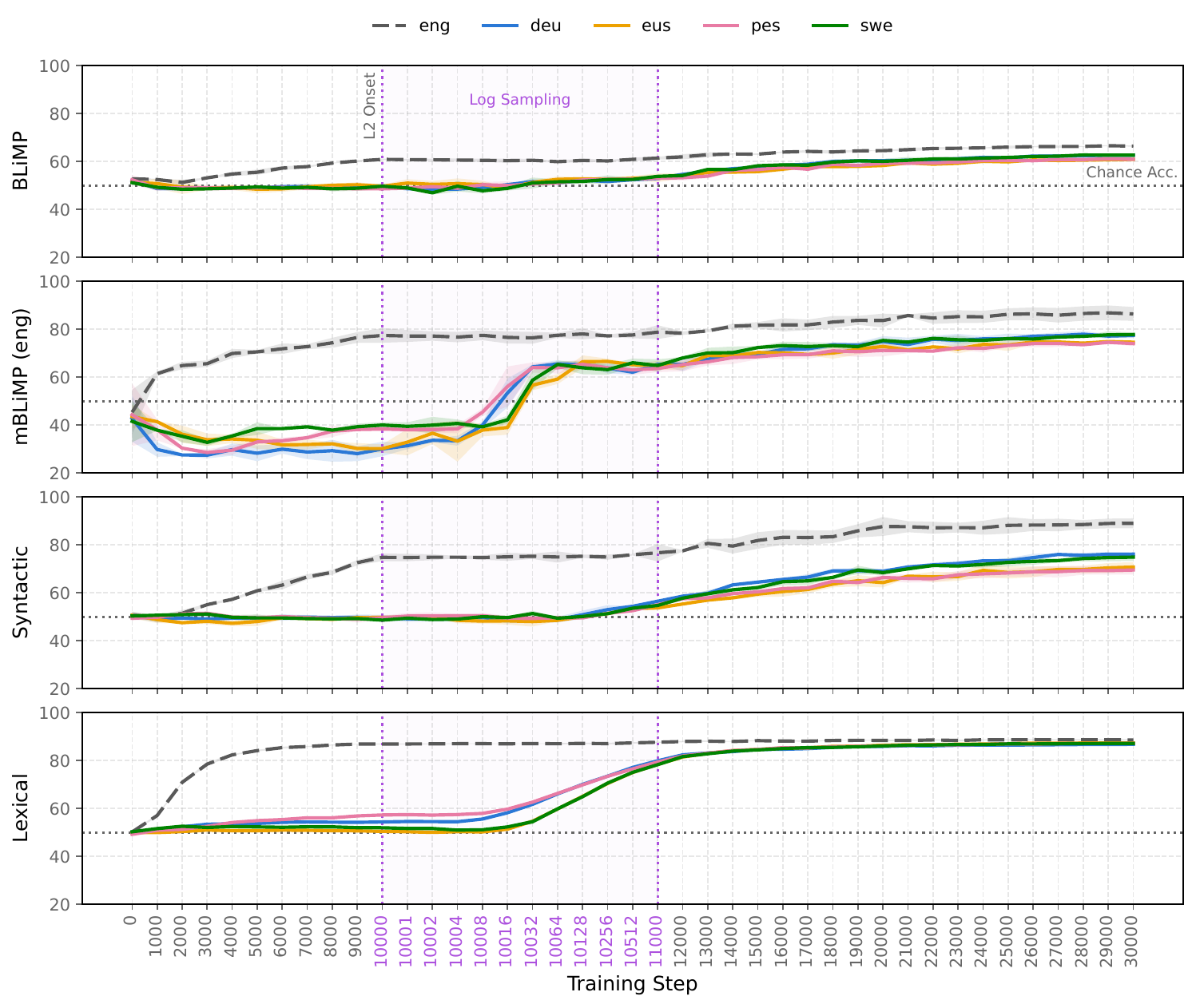}
    \caption{Evaluation accuracy over training for the bilingual \textbf{Phoneme LMs}. The first vertical dotted line marks the onset of joint L1--English training; horizontal dotted lines indicate chance performance. Lines correspond to average performance across 3 random seeds with $\pm$1 SD shading. For the first 1,000 steps after L2 introduction (10k--11k) we save logarithmically spaced checkpoints, highlighted with violet.}
    \label{fig:performance_over_training}
\end{figure*}

\paragraph{Cross-modal comparison}
Figure~\ref{fig:acc_dots} compares grammatical performance across L1s in the orthographic and phoneme conditions. The orthographic LMs achieve higher absolute scores overall, but the ordering of L1 conditions is considerably more consistent among the phoneme LMs, with German and Swedish outperforming Persian and Basque across grammatical evaluations. This is also visible in the correlation coefficients computed over L1 accuracy and syntactic distance. Specifically, phoneme LMs achieve correlations of $-0.94$, $-0.99$, and $-0.95$ for BLiMP, MultiBLiMP, and BabySLM Syntactic, respectively, compared to $-0.47$, $+0.13$ and $-0.46$ for orthographic LMs. However, due to the small number of L1s, these correlations are descriptive and should be interpreted cautiously rather than as evidence of a predictive relationship between syntactic distance and performance. Overall, models show stronger alignment between L1 identity and grammatical performance under phonemic than orthographic input.

\subsection{Training Dynamics}

Final-checkpoint scores can conceal differences in when and how linguistic knowledge emerges. We therefore evaluate intermediate checkpoints throughout training. Checkpoints are saved every 1,000 steps, with additional logarithmically spaced checkpoints immediately following the introduction of English. This allows us to observe the effect of the L1--L2 interaction on the learning trajectory. 

\begin{figure*}[ht!]
    \centering
    \includegraphics[width=\linewidth]{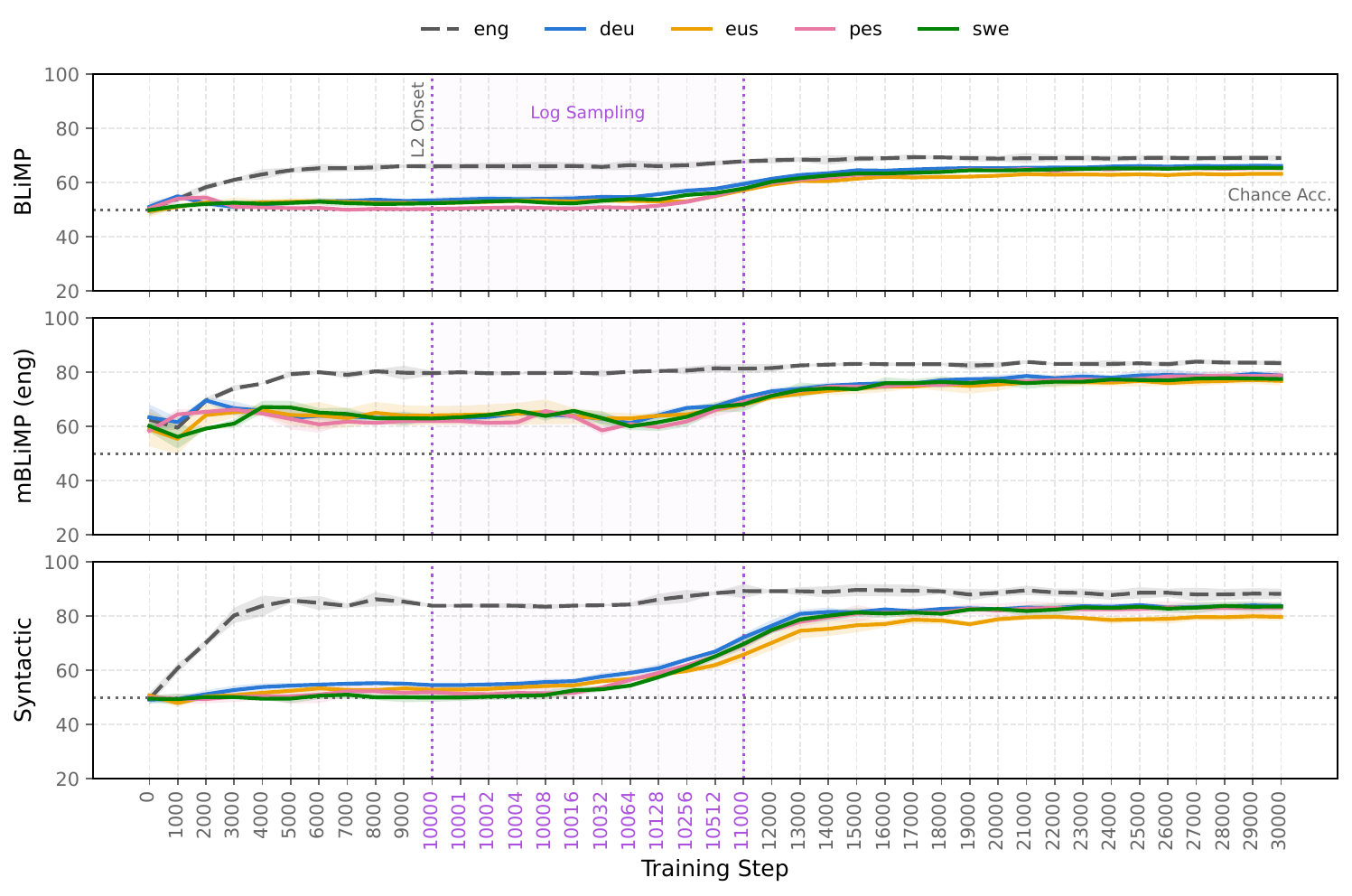}
    \caption{Evaluation accuracy over training for the bilingual \textbf{Orthographic LMs}.}
    \label{fig:orthographic}
\end{figure*}

Figure~\ref{fig:performance_over_training} presents phoneme LM trajectories for the four benchmarks: BLiMP, MultiBLiMP and BabySLM Syntactic and Lexical metrics. Corresponding trajectories for the orthographic LMs are reported in Figure~\ref{fig:orthographic}. In the orthographic condition, differences among L1 trajectories are smaller and more benchmark-dependent, with no persistent ordering that consistently follows syntactic distance. Detailed trajectories for each BLiMP sub-category of phenomena are included in Appendix~\ref{app:blimp}.

\paragraph{Grammatical learning}
In phoneme LMs, the BabySLM syntactic trajectories reveal a sustained separation among the L1 conditions after the introduction of English. The German model achieves the highest accuracy throughout much of joint training, while Swedish also generally remains above Persian and Basque. This ordering emerges over the course of English exposure and persists toward the end of training, suggesting that L1 identity is associated not only with final performance but also with the trajectory of English grammatical learning under phonemic input.
The separation is also present, but less pronounced, in the phonemic versions of BLiMP and English MultiBLiMP. Benchmark composition may contribute to these differences in sensitivity. BLiMP averages over a large and heterogeneous set of phenomena, potentially obscuring effects confined to particular categories, whereas MultiBLiMP evaluates only two subject--verb agreement phenomena. BabySLM Syntactic covers a smaller set of relatively simple contrasts, some involving word order, and may therefore expose L1-related differences that are diluted in broader aggregate scores. Dynamics for individual BLiMP phenomena are analyzed in Appendix~\ref{app:blimp}. 

More generally, bilingual-acquisition research suggests that cross-linguistic influence should be property-specific rather than a uniform consequence of global language similarity. In particular, influence has been associated with structural overlap in the relevant construction \cite{HulkMuller2000}. The stronger BabySLM separation may reflect overlap in particular word-order or syntactic patterns, rather than syntactic distance as a whole.

\paragraph{Lexical learning}
The BabySLM lexical trajectories reveal a transient pattern that is absent from the final scores. During L1-only training, the German and Persian models start with higher English lexical accuracy than the Swedish and Basque models. This similarity vanishes after the first 1,000 steps of L2 exposure, where all models rapidly converge to near 88\% accuracy.
German and Persian are the two selected languages with lower phoneme-inventory distance from English, making this ordering consistent with an effect of phoneme-inventory similarity in early lexical performance. Although the effect has a small duration, it is consistent with an L1-specific influence in accelerating the acquisition of a linguistic property that has been observed in studies with bilingual children \cite{Serratrice2013-tk}.

\section{Discussion}

Our results suggest that cross-linguistic influence varies across linguistic levels and stages of learning. Phoneme-inventory similarity is associated with an early lexical advantage that disappears after sustained English exposure, whereas syntactic similarity aligns with more persistent grammatical differences. This is consistent with accounts in which bilingual systems remain differentiated but interact selectively depending on specific properties \cite{ParadisGenesee1996,Serratrice2013-tk}. Importantly, the same pattern was substantially weaker and less consistent in orthographic text models.

The lexical trajectories further show that cross-linguistic influence may affect learning rate rather than final performance, highlighting the limits of final-checkpoint evaluation. Across modalities, orthographic input yields higher grammatical accuracy, whereas phonemic input produces clearer L1-related ordering. These interpretations remain exploratory, as each distance configuration is represented by a single language and the L1 corpora differ substantially in composition (Appendix~\ref{app:data_distribution}).

\section{Conclusion}

In this work, we investigated cross-linguistic effects in bilingual BabyLMs trained on orthographic and phonemic input. Orthographic LMs achieved higher overall grammatical accuracy, whereas phoneme LMs showed clearer L1-related variation: syntactic similarity aligned with more persistent  differences, while phoneme-inventory similarity aligned with an early lexical advantage that disappeared after English exposure. These findings suggest that input modality shapes both performance and cross-linguistic influence during learning.

\section*{Limitations}
Our work is limited in its cross-linguistic analysis by the selection of only four L1 languages (German, Swedish, Persian, Basque) and only one L2 (English). To strengthen evidence for cross-linguistic effects it is important to expand analysis to more typologically diverse pairs of languages. Additionally, we only utilized one model architecture (GPT-2) with hyperparameters from the Goldfish models \cite{chang-etal-2026-goldfish}. We train bilingual LMs using sequential-interleaved training, but other data mixing methods have been explored in the literature \cite{constantinescu-etal-2025-investigating, arnett-etal-2025-acquisition}, including pure sequential training (L1 followed by L2) or interleaved training (L1 mixed with L2). Future work should train phoneme LMs to disentangle the effect of these conditions. Lastly, grapheme LMs \cite{goriely-etal-2024-babble, bunzeck-etal-2024-graphemes} provide an interesting alternative to both tested configurations. Exploring whether character-based training aligns more closely with phoneme or orthographic LMs would help better identify the factors of the observed cross-linguistic differences.

Our goal when training phoneme LMs was to provide more ecological training conditions when simulating language acquisition. However, phonemes are only an intermediate step in getting to the real input that children are exposed to: speech. In this case, while it allows us to get closer to child input, phoneme input omits key features that are crucial in acquisition, such as prosody, timing, and stress. Cross-linguistic effects should also be investigated with raw speech input, following the quest to build a more accurate reconstruction of the (bilingual) child learner \cite{Dupoux_2018}.

\bibliography{acl_latex}

\appendix

\section{Training Details}
\label{app:training}

\begin{table}[h]
    \centering 
    \resizebox{\linewidth}{!}{
    \begin{tabular}{lr}
    \toprule
    Training Steps\\
    ~~Phase one (L1) & 10k\\
    ~~Phase two (L1+L2) & 20k\\
    ~~Monolingual & 30k\\
    Embedding params\\
    ~~(orthographic / phoneme)  & 23M / 500k\\
    Non-embedding params & 85M\\
    Total & 108M / 85M\\
    Layers & 12\\
    Embedding size & 768\\
    Intermediate hidden & 3072\\
    Attention heads & 12\\
    Attention head size & 64\\
    Learning rate & 1e-4\\
    Batch size& 32\\
    Epochs per language &  10\\
    Activation function & GELU\\
    Max sequence length & 512\\
    Position embedding & Absolute\\
    Learning rate decay & Linear\\
    Warmup steps & 10\% of pretraining\\
    Adam $\epsilon$ & 1e-6\\
    Adam $\beta_1$ & 0.9\\
    Adam $\beta_2$ & 0.999\\
    Dropout & 0.1\\
    Attention dropout & 0.1\\
    \bottomrule
    \end{tabular}
    }
    \caption{Training hyperparameters.}
    \label{tab:hyperparameters}
\end{table}

\begin{figure}[h]
    \centering
    \includegraphics[width=\linewidth]{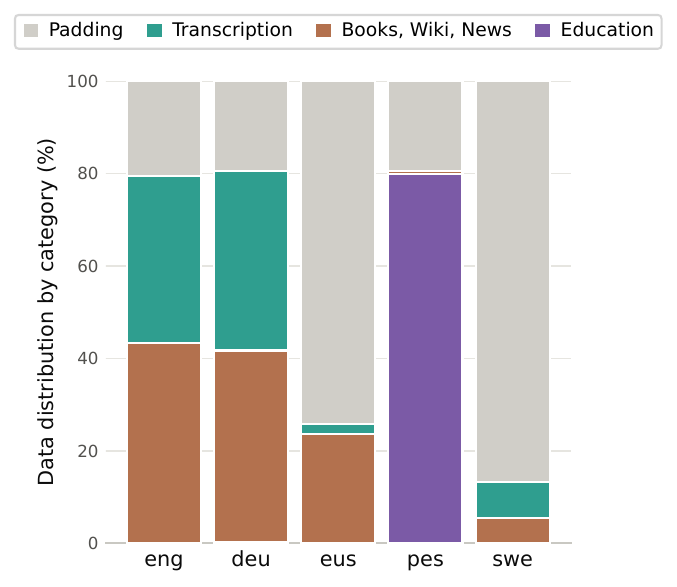}
    \caption{Data category distribution across languages.}
    \label{fig:data_distribution}
\end{figure}

\begin{figure*}[t!]
        \centering
    \includegraphics[width=0.9\linewidth]{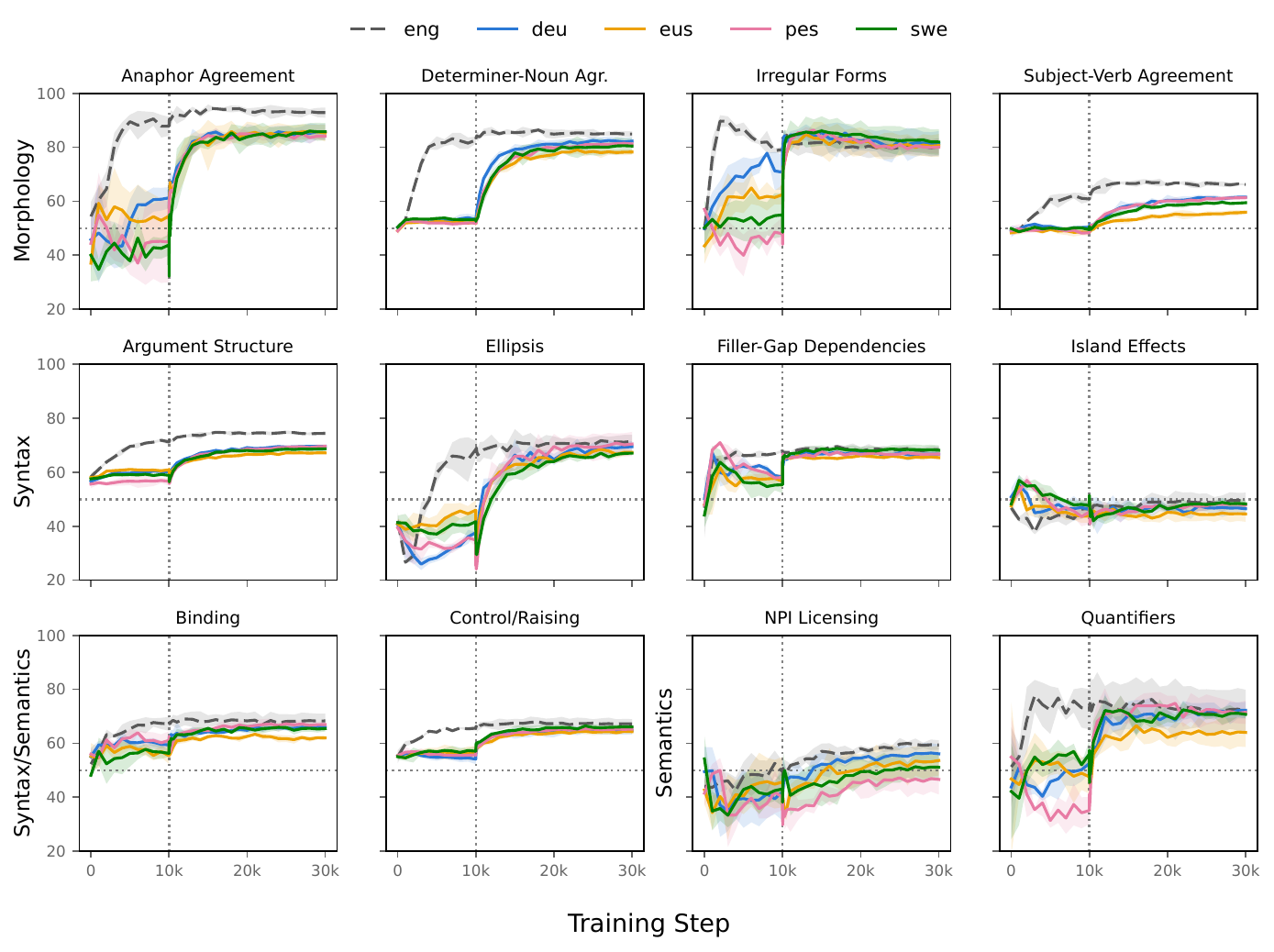}
    \caption{BLiMP accuracy per category for \textbf{Orthographic LMs} throughout training. Dashed lines indicate L2 onset and random change accuracy (50\%). Lines correspond to average performance across 3 random seeds with $\pm$1 SD shading. Plots in each row are grouped according to the broader linguistic field: Morphology (first row), Syntax (second row), Syntax/Semantics (third row left), Semantics (third row right).}
    \label{fig:detailed_blimp_text}
\end{figure*}

\begin{figure*}[t!]
        \centering
    \includegraphics[width=0.9\linewidth]{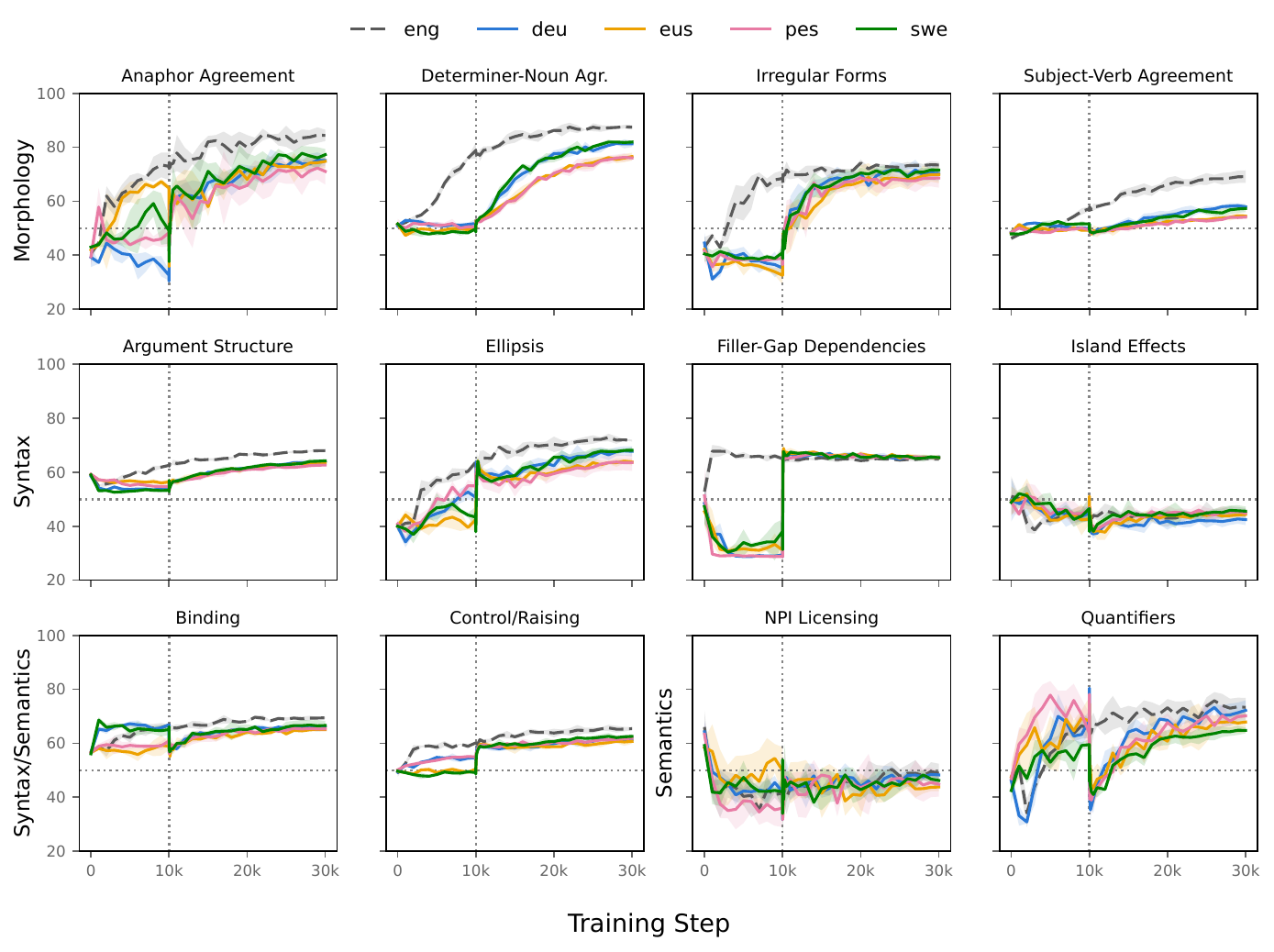}
    \caption{BLiMP accuracy per category for \textbf{Phoneme LMs} throughout training. }
    \label{fig:detailed_blimp_phone}
\end{figure*}

We use training hyperparameters from the GoldFish \cite{chang-etal-2026-goldfish} suite of models with minimal changes, which we include in Table~\ref{tab:hyperparameters}. Orthographic LMs are trained with a BPE tokenizer and a vocabulary of 30,000 tokens. For phoneme LMs we construct a phoneme-based tokenizer consisting of phonemes with more than 10 occurrences in the bilingual corpus. A separate tokenizer is trained for each language pair on the bilingual training data. For both training phases (sequential, interleaved) we keep the same optimizer and learning rate scheduler, changing only the data source. Training was done with mixed-precision (bf16) and took approximately 1 hour on \texttt{1x RTX 5090 GPU} for each model in both orthographic and phoneme conditions. In total, for 3 different seeds, this amounts to 30 hours of compute across both modalities.

Differences between modalities are mainly attributed to the tokenizer, which accounts for diverging embedding sizes (23M vs. 500k), and a reduced effective context in phoneme LMs. Even though all models are trained with a sequence length of 512 tokens, the uncompressed phoneme stream allows less content to fit in one block compared to BPE.

\section{Detailed BLiMP Performance}
\label{app:blimp}

\begin{table}[h]
\resizebox{\linewidth}{!}{
\begin{tabular}{ll|cccc}
\toprule
 & \textbf{L1} & Morphology & Syntax & Syntax/Semantics & Semantics \\
\midrule\addlinespace[2.5pt]
\multirow{5}{*}{\rotatebox{90}{\textbf{Orthographic}}} & eng & \cellcolor{heat!54} 79.0 & \cellcolor{heat!25} 64.9 & \cellcolor{heat!31} 67.9 & \cellcolor{heat!23} 64.0 \\
 & deu & \cellcolor{heat!47} 75.6 & \cellcolor{heat!19} 61.7 & \cellcolor{heat!26} 65.4 & \cellcolor{heat!19} 62.0 \\
 & eus & \cellcolor{heat!40} 71.9 & \cellcolor{heat!15} 59.8 & \cellcolor{heat!21} 63.0 & \cellcolor{heat!10} 57.4 \\
 & pes & \cellcolor{heat!46} 74.8 & \cellcolor{heat!20} 62.2 & \cellcolor{heat!28} 66.2 & \cellcolor{heat!6} 55.7 \\
 & swe & \cellcolor{heat!44} 74.2 & \cellcolor{heat!19} 62.1 & \cellcolor{heat!27} 65.7 & \cellcolor{heat!12} 58.3 \\
\midrule
\multirow{5}{*}{\rotatebox{90}{\textbf{Phoneme}}} & eng & \cellcolor{heat!55} 79.4 & \cellcolor{heat!16} 60.4 & \cellcolor{heat!31} 67.8 & \cellcolor{heat!11} 57.8 \\
 & deu & \cellcolor{heat!39} 71.6 & \cellcolor{heat!11} 58.1 & \cellcolor{heat!24} 64.2 & \cellcolor{heat!9} 57.0 \\
 & eus & \cellcolor{heat!32} 68.3 & \cellcolor{heat!11} 58.1 & \cellcolor{heat!22} 63.2 & \cellcolor{heat!0} 52.6 \\
 & pes & \cellcolor{heat!30} 67.4 & \cellcolor{heat!11} 58.0 & \cellcolor{heat!23} 63.9 & \cellcolor{heat!3} 54.0 \\
 & swe & \cellcolor{heat!40} 72.1 & \cellcolor{heat!13} 59.1 & \cellcolor{heat!25} 64.9 & \cellcolor{heat!1} 53.0 \\
\bottomrule
\end{tabular}

}
\caption{BLiMP accuracy across linguistic fields.}
\label{tab:blimp_detailed}
\end{table}

Aggregate benchmark performance may hide L1 or modality effects on the level of specific linguistic phenomena. To investigate this, we plot learning
trajectories for every BLiMP category (Figures \ref{fig:detailed_blimp_text}, \ref{fig:detailed_blimp_phone}) and average performance for each linguistic field in Table \ref{tab:blimp_detailed}. Overall, models followed different learning patterns for each phenomenon. Performance either saturated early (Filler-gap Dependencies, Control/Raising), improved throughout training (Determiner-Noun Agr., Ellipsis), or stayed near chance (Island Effects, NPI Licensing). As expected, English monolingual LMs generally performed better than bilingual for both modalities, particularly in morphology. Overall, models score highest in morphology and lowest in semantics.

Orthographic LMs showed no strong cross-linguistic effects (Fig. \ref{fig:detailed_blimp_text}), except in the case of \texttt{eus} were we occasionally see reduced performance (S-V Agr., Quantifiers). In contrast, for phoneme LMs, we notice that the languages most syntactically similar to English (\texttt{deu}, \texttt{swe}) scored higher for specific phenomena, two in morphology (Det-Noun Agr., S-V Agr.) and one in syntax (Ellipsis), indicating an influence of syntactic or genealogical similarity on bilingual model performance that was also observed for the aggregate BLiMP accuracy.

When comparing across modalities, phoneme LMs achieve similar scores to the orthographic ones, except for semantics. This might stem from the reduced context length of phoneme LMs, which makes it harder to learn long-range semantic dependencies. Training stability also depended on modality, with phoneme LMs taking longer to converge for some phenomena (Irregular Forms, Anaphor Agr., Quantifiers) compared to orthographic ones.

\section{Data Distribution}
\label{app:data_distribution}
Data distribution for all languages is illustrated in Figure \ref{fig:data_distribution}. Languages in the 100M word tier (German, Persian, English) were downsampled to 10M equivalent English words, adjusted using byte-premiums \cite{arnett-etal-2024-bit}, and approximating as much as possible the data distribution of English. Data source --- transcribed spontaneous speech or written language --- can influence model performance, particularly for phoneme LMs where training data must undergo phonemization. While an important factor, we opted for developmentally plausible corpora rather than parallel or category-matched training datasets. The influence of data-source on cross-linguistic effects should be investigated in future work.

\end{document}